# PPL: Point Cloud Supervised Proprioceptive Locomotion Reinforcement Learning for Legged Robots in Crawl Spaces

Bida Ma[1], Nuo Xu[1], Chenkun Qi[1], *Senior Member, IEEE*, Xin Liu[1], Yule Mo[1], Jinkai Wang[1], and Chunpeng Lu[2].

***Abstract*—Legged locomotion in constrained spaces (called crawl spaces) is challenging. In crawl spaces, current proprioceptive locomotion learning methods are difficult to achieve traverse because only ground features are inferred. In this study, a point cloud supervised RL framework for proprioceptive locomotion in crawl spaces is proposed. A state estimation network is designed to estimate the robot's collision states as well as ground and spatial features for locomotion. A point cloud feature extraction method is proposed to supervise the state estimation network. The method uses representation of the point cloud in polar coordinate frame and MLPs for efficient feature extraction. Experiments demonstrate that, compared with existing methods, our method exhibits faster iteration time in the training and more agile locomotion in crawl spaces. This study enhances the ability of legged robots to traverse constrained spaces without requiring exteroceptive sensors.**

## I. INTRODUCTION

IN recent years, legged robots have demonstrated remarkable terrain traversal capabilities, exhibiting significant application value. Reinforcement learning (RL) enables legged robots to generate more flexible actions in complex terrains than model based control, consequently improving their adaptability in challenging environments [1][2][3][4]. In many applications, such as earthquake and mine's search and rescue, the working environments have constrained spaces including low-ceiling tunnels and small caves (called as crawl spaces). In these cases, the traversability of legged robots faces new challenges. Learning based locomotion using exteroceptive sensors such as LiDAR or depth cameras can generate flexible actions in complex terrains [5][6][7][8]. However, these exteroceptive sensors could have large noises, making the exteroceptive locomotion sensitive to environmental conditions. Learning based locomotion using proprioceptive sensors including joint encoders and IMU is more robust to these environmental conditions. Therefore, this study is focused on the proprioceptive locomotion reinforcement learning for legged robots in crawl spaces.

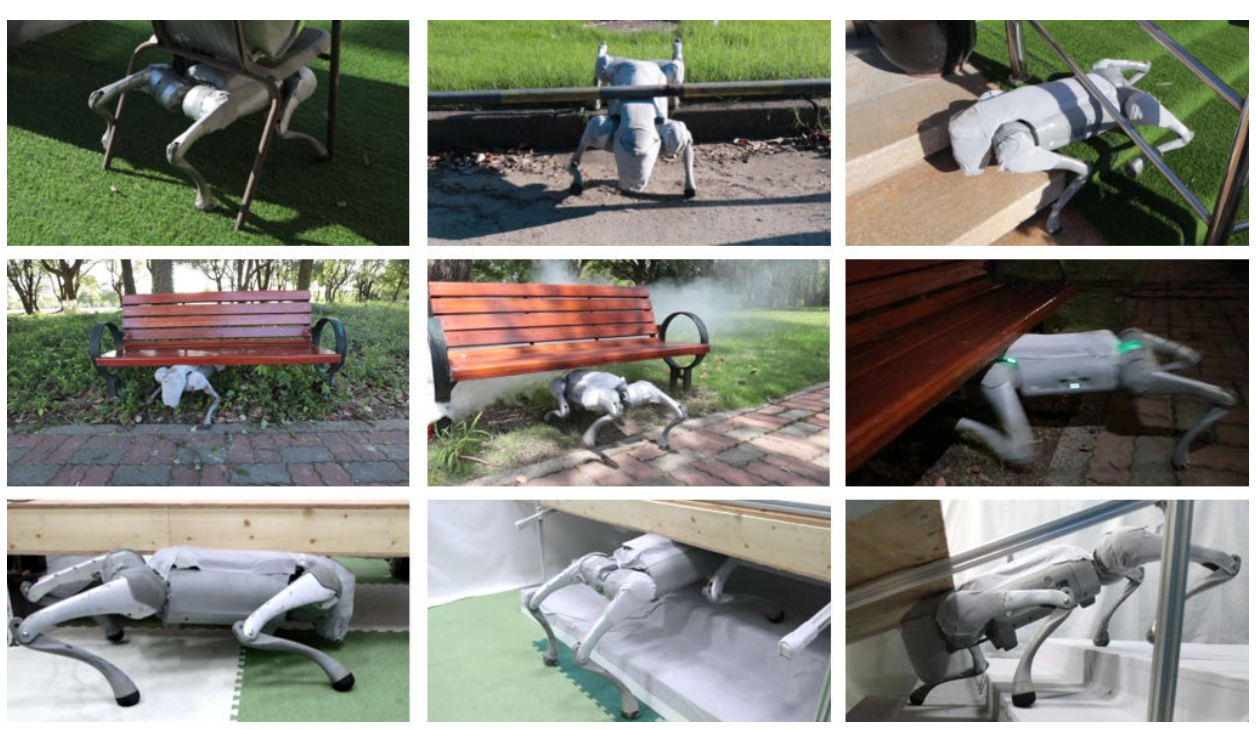

Figure 1. We deploy the policy on the real robot. Extensive experiments demonstrate the effectiveness of the proposed method in crawl spaces, and the experimental video is provided in the supplemental material.

Currently, learning based locomotion using proprioceptive sensors mainly consider open spaces [9][10][11], where there is no ceiling or the ceiling is high enough. These methods cannot make the robot traverse through crawl spaces, because the elevation map used as privileged terrain information for the locomotion learning cannot describe spatial structures such as tunnels and caves. Instead of using elevation map as privileged terrain information, a collision domain concept (a bounding box of the robot that can be penetrable with environments) is proposed in [12] to describe spatial structures such as tunnels and caves. However, the collision domain has environmental inner information which is redundant for the locomotion learning. Taking depth images as privileged terrain information is another method [13]. However, to infer all spatial structures around robots for omnidirectional locomotion, multiple depth images are needed, where texture rendering requires excessive computation. Using point cloud as terrain information could speed up locomotion training, because texture rendering is not required [14]. Works such as [15] on feature encoding for point clouds in Cartesian coordinates, typically employ sophisticated networks that demand training with large-scale training datasets. Due to the lack of large-scale training datasets in legged robots locomotion, there are significant challenges for the use in legged robots.

In this paper, to traverse through crawl spaces, we propose a Point cloud supervised Proprioceptive Locomotion (PPL) reinforcement learning method. The method achieves

The supplement video is available at https://www.bilibili.com/video/BV1gmbvzuEuW/

1 School of Mechanical Engineering, Shanghai Jiao Tong University, Shanghai, China

2 Lenovo Corporation, Shanghai, China

* Chenkun Qi (chenkqi@sjtu.edu.cn) is the corresponding author.

proprioceptive locomotion for legged robots in crawl spaces. As shown in Fig. 1, real-world experiments on UnitreeGo2 demonstrate the effectiveness of the method.

The main contributions of this work are as follows:

- Point-Cloud Supervision Framework: A point cloud supervised RL framework for proprioceptive locomotion in crawl spaces is proposed. This framework uses 3D representations (point clouds) as privileged information solely during training and does not use point clouds during deployment.
- Feature Extraction Method: A point cloud feature extraction method is proposed to supervise the state estimation network. The method is novel on representation of the point cloud in polar coordinate frame and MLPs for efficient feature extraction.
- Real-World Validation: Experiments in flat tunnel, stairs tunnel and outdoors are conducted to validate the effectiveness of the proposed method.

## II. Related Works

Works with exteroceptive sensors [5][16][17][18][19] have enhanced the traversability for legged robots in crawl spaces. However, exteroceptive sensors are sensitive to environmental conditions, often have large noises, and even fail. Proprioception-based reinforcement learning for locomotion control in legged robots [2][20][21] is more robust to environmental conditions. However, proprioceptive locomotion is mainly studied in open spaces, the problem for crawl spaces is not solved.

Currently, there are only a few studies [12][13] on proprioceptive locomotion in crawl spaces. Work [12] characterizes spatial structures by quantifying the penetration depth of the collision domain into environments, and develops a collision estimator to estimate collision states of different body parts. However, in rugged-terrain caves or stairs tunnels, a portion of the collision domain persistently penetrates into objects, resulting in a redundant representation. Work [13] uses multidirectional depth images as privileged environmental information, and incorporates domain randomization to address visual occlusion scenarios. But the acquisition and processing of multidirectional depth images during the training require heavy computation and long time.

Point clouds can provide omnidirectional environmental information. A fast and efficient point cloud feature extraction method is required for critic to accurately evaluate robot state value. In other fields [15], complex neural networks are used to handle point cloud's structural disorder and size variations. However, large-scale datasets are required for pre-training, and heavy computation is also a problem. In exteroceptive locomotion works [6][7][14][22][23], point clouds or voxel grids are typically represented in Cartesian coordinate frame and processed using complex neural networks for feature extraction. These methods require substantial computational resources during training.

According to 3D representations during training and deployment, the existing frameworks for locomotion can be classified into two types. The first type of framework is that the framework relies on 3D representations neither during training nor during deployment [2][9][20]. It can't be applied for locomotion in crawl spaces. The second type of framework is that the framework relies on 3D representations during both training and deployment [6][7][14][22][23]. It can be applied for locomotion in crawl spaces by leveraging external perception.

Unlike prior works, our method uses 3D representation of the point cloud in polar coordinate frame for training, and does not require 3D representation during deployment. Moreover, MLPs rather than complex neural networks are used for efficient feature extraction of the point cloud.

## III. Method

To make legged robots traverse through crawl spaces, we propose a point cloud supervised proprioceptive locomotion reinforcement learning method. Our approach overcomes the limitation that elevation map cannot describe spatial structures in crawl spaces. Here, we first introduce our method including network architecture, PPL-Net and reward design, and then provide training details.

### A. Network Architecture

As shown in Fig. 2, our framework comprises three components: actor, critic, and PPL-Net. The PPL-Net will be introduced in Section III-B. The actor network receives current proprioceptive information, control commands, and estimated state information to generate action. The critic network receives robot states to generate state-value estimate. Both the actor and critic are implemented as multi-layer perceptions (MLPs). The reward functions are used to calculate the rewards after an action is applied. The accumulated rewards in a period of time are used to supervise the training of critic. The critic is then used to help to compute the policy gradient for training the actor. To avoid the policy performance degradation of traditional multi-stage learning methods, here a one-stage asymmetric actor-critic architecture is employed in the policy training [24]. Proximal Policy Optimization (PPO) [25] method is used in the network parameters optimization.

*1) Actor Network:* The actor network's inputs at time $t$ include control commands $v_x^{cmd} \in \mathbb{R}$, $v_y^{cmd} \in \mathbb{R}$, and $\omega_z^{cmd} \in \mathbb{R}$, which represent desired longitudinal, lateral, and yaw velocities respectively in the robot base frame, and current proprioceptive information $\boldsymbol{o}_t^p \in \mathbb{R}^{42}$. The proprioceptive information $\boldsymbol{o}_t^p$ at time $t$ is defined as

$$\boldsymbol{o}_t^p = [\boldsymbol{\omega}_t \quad \boldsymbol{g}_t \quad \boldsymbol{\theta}_t \quad \dot{\boldsymbol{\theta}}_t \quad \boldsymbol{a}_{t-1}] \tag{1}$$

which consists of the base angular velocity $\boldsymbol{\omega}_t \in \mathbb{R}^3$ in the robot base frame, gravity vector projection $\boldsymbol{g}_t \in \mathbb{R}^3$ in the robot base frame, joint positions $\boldsymbol{\theta}_t \in \mathbb{R}^{12}$, joint velocities $\dot{\boldsymbol{\theta}}_t \in \mathbb{R}^{12}$, and action from the last time step $\boldsymbol{a}_{t-1} \in \mathbb{R}^{12}$.

To make the actor generate more appropriate action, actor's inputs incorporate estimated state information, which are base linear velocity $\tilde{\boldsymbol{v}}_t \in \mathbb{R}^3$ in the robot base frame, head, base and hips collision states $\tilde{\boldsymbol{c}}_t \in \mathbb{R}^6$, and ground and spatial latent representation $\tilde{\boldsymbol{z}}_t^l \in \mathbb{R}^{20}$.

The action $\boldsymbol{a}_t \in \mathbb{R}^{12}$ is defined as offsets from the joint default positions $\boldsymbol{\theta}_{default} \in \mathbb{R}^{12}$, and the relationship between action

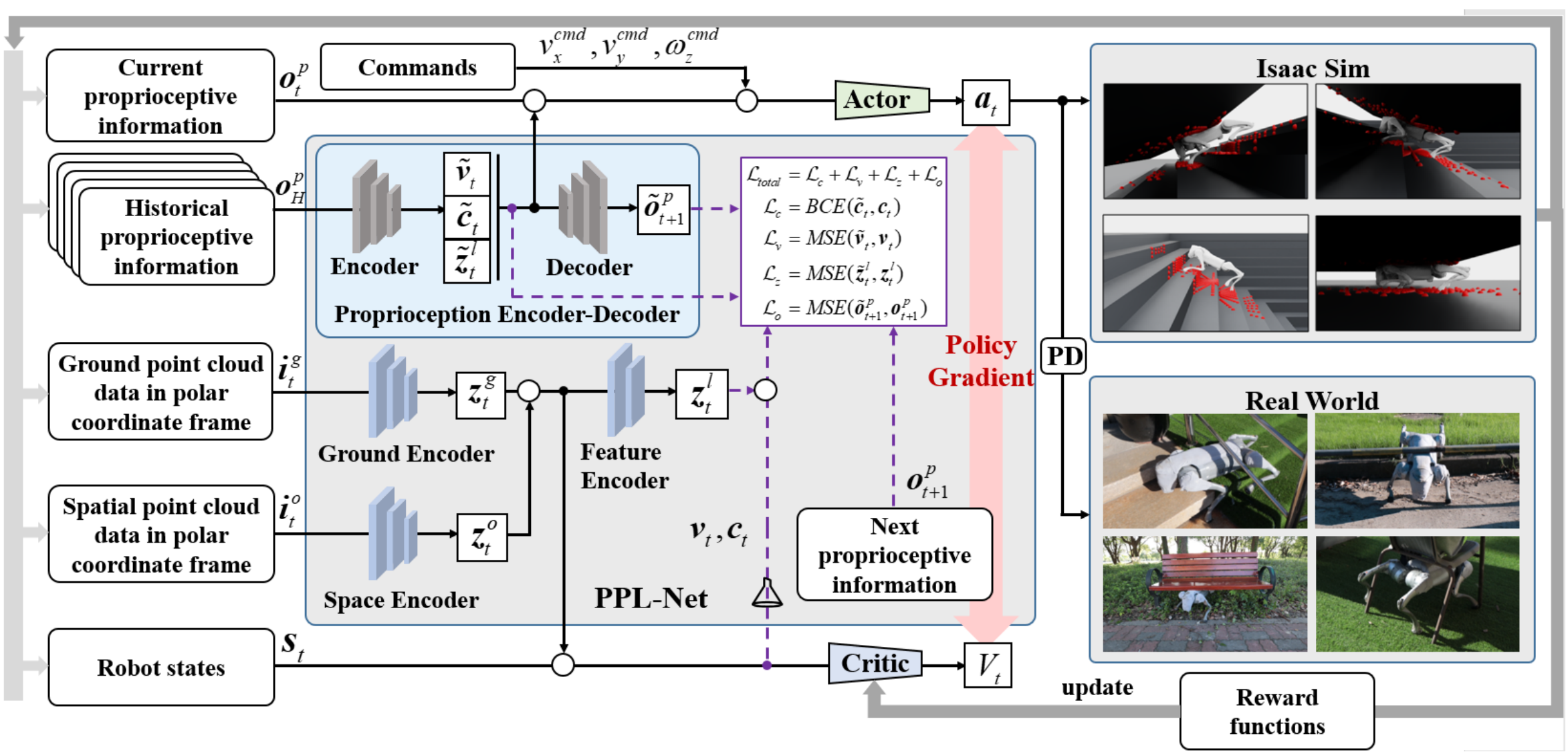

Figure 2. Overview of the proposed learning method. It comprises three components: actor, critic, and PPL-Net. The proposed PPL-Net includes proprioception Encoder-Decoder, Ground Encoder, Space Encoder and Feature Encoder. The proprioception Encoder-Decoder is designed to infer ground and spatial features as well as collision states from proprioceptive sensors' data. The Ground Encoder, Space Encoder and Feature Encoder are designed to get the true ground and spatial features from point cloud data in the polar coordinate frame. The one-stage asymmetric actor-critic architecture is employed in the policy training.

and joint position commands $\boldsymbol{\theta}_{target} \in \mathbb{R}^{12}$ is

$$\boldsymbol{\theta}_{target} = \boldsymbol{a}_t + \boldsymbol{\theta}_{default} \tag{2}$$

*2) Critic Network:* The critic network receives robot states to generate state-value estimate. In this work, robot states $\boldsymbol{s}_t$ include current proprioceptive information $\boldsymbol{o}_t^p$ and privileged information, which are base linear velocity $\boldsymbol{v}_t \in \mathbb{R}^3$ in the robot base frame, head, base and hips collision states $\boldsymbol{c}_t \in \mathbb{R}^6$, base height $\boldsymbol{h}_t \in \mathbb{R}$, foot contact forces $\boldsymbol{f}_{cf} \in \mathbb{R}^{12}$ in the world frame, external force $\boldsymbol{f}_{ef} \in \mathbb{R}^3$ in the world frame, external force position $\boldsymbol{p}_{ef} \in \mathbb{R}^3$ in the robot base frame, and point cloud encoders' outputs $\boldsymbol{z}_t^g \in \mathbb{R}^{187}$ and $\boldsymbol{z}_t^o \in \mathbb{R}^{90}$. The dimensions of $\boldsymbol{z}_t^g$ and $\boldsymbol{z}_t^o$ are set by empirical adjustments. Each dimension of $\boldsymbol{c}_t$ is a binary variable, where 0 represents no collision and 1 represents collision. The robot states can be represented as

$$\boldsymbol{s}_t = [\boldsymbol{o}_t^p \quad \boldsymbol{v}_t \quad \boldsymbol{c}_t \quad \boldsymbol{h}_t \quad \boldsymbol{f}_{cf} \quad \boldsymbol{f}_{ef} \quad \boldsymbol{p}_{ef} \quad \boldsymbol{z}_t^g \quad \boldsymbol{z}_t^o] \tag{3}$$

## B. PPL-Net

Current proprioceptive locomotion works [2][9][10][11][26] utilize proprioceptive information to infer ground features, however, locomotion in crawl spaces needs not only ground features but also spatial features, such as tunnel ceiling positions. In exteroceptive locomotion works [5][6][7][27][28], point clouds or depth images are employed to represent ground and spatial information of environments. Compared with depth images, employing point cloud based environmental information has advantage of efficient computation during the training phase [14]. Therefore, in our PPL method, we propose a Point cloud supervised Proprioceptive Locomotion Network (PPL-Net), which utilizes point cloud supervised learning to infer ground and spatial information and collision states in crawl spaces.

*1) PPL-Net Architecture:* The PPL-Net consists of one Proprioception Encoder-Decoder network, two privileged environmental information encoders (Ground Encoder and Space Encoder) and one Feature Encoder as shown in Fig. 2. The Proprioception Encoder processes historical proprioceptive information $\boldsymbol{o}_H^p = [\boldsymbol{o}_{t-4}^p \ \ldots \ \boldsymbol{o}_{t-1}^p \ \boldsymbol{o}_t^p] \in \mathbb{R}^{210}$ to generate estimations of base linear velocity $\tilde{\boldsymbol{v}}_t$, head, base and hips collision states $\tilde{\boldsymbol{c}}_t$ as well as ground and spatial latent representation $\tilde{\boldsymbol{z}}_t^l$. Here we use the five measurements of most recent proprioceptive information as the work [20]. The Proprioception Decoder reconstructs the next time step proprioceptive information $\tilde{\boldsymbol{o}}_{t+1}^p$ utilizing estimations $[\tilde{\boldsymbol{v}}_t \ \tilde{\boldsymbol{c}}_t \ \tilde{\boldsymbol{z}}_t^l]$. Ground Encoder and Space Encoder receive preprocessed ground point cloud data $\boldsymbol{i}_t^g \in \mathbb{R}^{720}$ and spatial point cloud data $\boldsymbol{i}_t^o \in \mathbb{R}^{720}$ respectively, to generate ground point cloud representation $\boldsymbol{z}_t^g$ and spatial point cloud representation $\boldsymbol{z}_t^o$ respectively. The Feature Encoder re-encodes ground point cloud representation $\boldsymbol{z}_t^g$ and spatial point cloud representation $\boldsymbol{z}_t^o$ to generate ground and spatial latent representation $\boldsymbol{z}_t^l \in \mathbb{R}^{20}$.

*2) Polar Coordinate Transformation of Point Clouds:* Extracting point cloud features usually uses complex neural networks which typically require prolonged training. The other problem is that the size of point cloud data varies temporally, which makes MLPs inapplicable. To address these issues, we propose a point cloud preprocessing method as shown in Fig. 3, which allows using MLPs for feature extraction to reduce training time. The point cloud data is generated through LiDAR pattern mode in the simulator, where the laser is uniformly distributed across all directions. Additionally, points beyond the laser's max measurement range do not affect the policy action. Therefore, to ensure that point cloud data maintains a fixed size, points beyond the laser's maximum measurement range are treated as being at the maximum measurement distance. Here we do not employ the Cartesian coordinate frame but instead represent the point cloud in the polar coordinate frame. The reason is that the radial distance is a single-valued function of the polar angles, and therefore, the point cloud data can be presented using a fixed order sequence. This data structure allows point's direction information to be implicitly embedded within the data sequence. The index of the point sequence denotes the polar angles, and the value of the elements denotes the radial distance. This processing can present the 3D point cloud as a 1D vector. For a point in 3D environment, its polar coordinate representation in the robot base frame is

$$\boldsymbol{\alpha} = \arctan(\sqrt{\boldsymbol{x}^2 + \boldsymbol{y}^2} \,/\, z) \tag{4}$$

$$\boldsymbol{\beta} = \arctan(\boldsymbol{y} \,/\, \boldsymbol{x}) \tag{5}$$

$$\boldsymbol{i} = \sqrt{\boldsymbol{x}^2 + \boldsymbol{y}^2 + z^2} \tag{6}$$

where $[\boldsymbol{x}\ \boldsymbol{y}\ z]$ is the Cartesian coordinate in the robot base frame, and $[\boldsymbol{\alpha}\ \boldsymbol{\beta}\ \boldsymbol{i}]$ is the polar coordinate in the robot base frame. Each combination of $\boldsymbol{\alpha}$ and $\boldsymbol{\beta}$ angles corresponds to a specific radial distance $\boldsymbol{i}$ . Specifically, the radial distance is the only input for point cloud encoder.

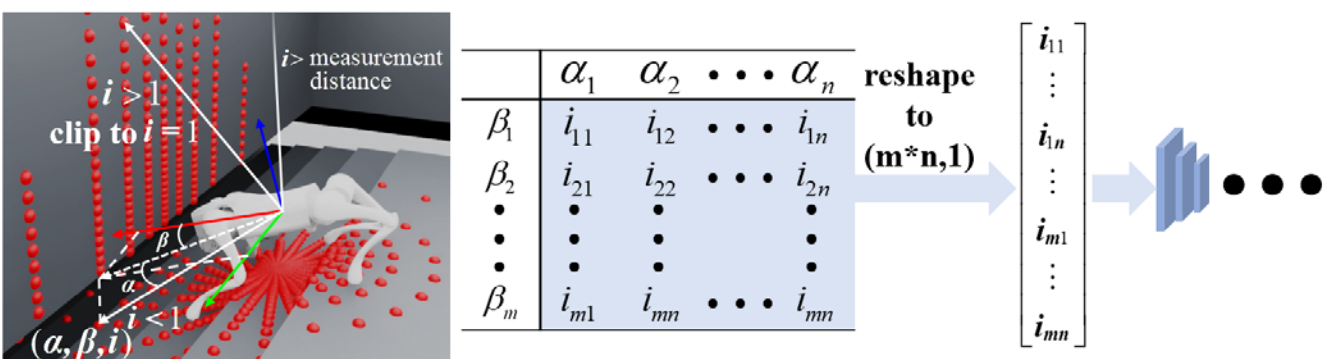

Figure 3. The point cloud preprocessing method. We represent the point cloud in the polar coordinate frame, and points beyond the laser's maximum measurement range are treated as being at the maximum measurement distance. Thus, the 3D point cloud can be presented as a 1D vector. Radial distances where $i > 1m$ are clipped to $i = 1m$ .

Since environmental information near the robot is more relevant to locomotion than distant environmental information, we retain only the point cloud near the robot. Here we use point cloud data within a 1m-radius ball centered on the robot base center. For points with $\boldsymbol{i} > 1m$ , the radial distance is clipped to $\boldsymbol{i} = 1m$ . Because the point cloud data is presented as a fixed-size of 1D vector, MLPs can be used for extracting the features of point cloud efficiently. In this work, to balance training speed and policy performance, $\boldsymbol{\alpha}$ for ground point clouds is discretized from -90° to 0° with 3° angular resolution, while $\boldsymbol{\alpha}$ for spatial point clouds spans 0° to 90° at the same resolution. $\boldsymbol{\beta}$ for both ground and spatial point clouds spans the full 360° azimuthal range, discretized at 15° increments. Therefore, the dimensions of the preprocessed ground and spatial point cloud data are $\boldsymbol{i}_t^g \in \mathbb{R}^{720}$ and $\boldsymbol{i}_t^o \in \mathbb{R}^{720}$ respectively.

3) *Loss Function of Proprioception Encoder-Decoder*: To ensure the accuracy of the estimation of proprioception Encoder-Decoder, supervised learning is used. The total loss function comprises four components, which are collision states estimation loss $\mathcal{L}_c = BCE(\tilde{\boldsymbol{c}}_t, \boldsymbol{c}_t)$ , velocity estimation loss $\mathcal{L}_v = MSE(\tilde{\boldsymbol{v}}_t, \boldsymbol{v}_t)$ , ground and spatial latent representation estimation loss $\mathcal{L}_z = MSE(\tilde{\boldsymbol{z}}_t^l, \boldsymbol{z}_t^l)$ , and next time step proprioceptive information estimation loss $\mathcal{L}_o = MSE(\tilde{\boldsymbol{o}}_{t+1}^p, \boldsymbol{o}_{t+1}^p)$ . Here MSE denotes Mean Squared Error. Therefore, the loss function of proprioception Encoder-Decoder is formulated as

$$\mathcal{L}_{total} = \mathcal{L}_c + \mathcal{L}_v + \mathcal{L}_z + \mathcal{L}_o \tag{7}$$

The same weights are used because the loss components have the same scale in our study. To generate active sensing motion for better estimation, the actor network receives gradients through $\tilde{\boldsymbol{v}}_t$ , $\tilde{\boldsymbol{c}}_t$ and $\tilde{z}_t^l$ from the PPL-Net losses, and the estimator losses are not detached as in [29].

## C. *Reward Design*

We design a group of rewards, including task reward, collision penalties, post-collision velocity reward and regularization rewards. All reward functions are listed in Table I. Note that the weight and other hyper-parameters are referred from current works [2][9][10][11][26] and refined empirically.

*1) Task Reward:* Similar to many locomotion works [2][9][10][11][26], the task reward term in our work is velocity command tracking reward including *xy* linear velocity tracking reward and yaw angle velocity tracking reward in the robot base frame.

*2) Whole-Body Collision Penalties:* Whole-body collision penalties are employed to encourage the robot to reduce undesired contact in crawl spaces. The whole-body collision penalties include penalties for leg collisions and body collisions. Leg collisions including foot collisions, thigh collisions and shank collisions typically occur against the ground. These collisions can be handled by adjusting joint positions in the corresponding leg. These behaviors can be easily learned by the policy without requiring complex force related penalty functions. Therefore, the leg collisions are penalized by the following step function

$$r_{collision_leg} = \sum_4 (c_{foot} + c_{thigh} + c_{shank}) \tag{8}$$

where $c_{foot}$ , $c_{thigh}$ and $c_{shank}$ represent the collision states of the robot's foot, thigh and shank respectively. A state value of 1 indicates collision occurrence, while 0 indicates no collision.

Body collisions including head collisions, base collisions and hip collisions typically occur against the crawl spaces' ceiling. These collisions should be handled by adjusting all joint positions in four legs. In our experience, this adjusting behavior is hard to be learned by the policy with only penalties of collision states. This is because the collision states penalties are sparse, and twelve joint actions are not easy to be searched simultaneously along the proper directions in the training. Therefore, we add force related components to densify collision penalties. In crawl spaces, horizontal component of the contact forces on the robot's head, base, and hips can prevent locomotion, thus, we add the horizontal component penalties of the contact force for robot's head, base and hips.

$$\begin{aligned} r_{collision_body} = {} & \lambda_1(1 - \exp(-\mu_1 \left\| \boldsymbol{f}_{head}^{xy} \right\|) + \lambda_2(1 - \exp(-\mu_2 \left\| \boldsymbol{f}_{base}^{xy} \right\|)) \\ & + \lambda_3(1 - \exp(-\mu_3 \sum_4 \left\| \boldsymbol{f}_{hip}^{xy} \right\|)) + c_{head} + c_{base} + \sum_4 c_{hip} \end{aligned} \tag{9}$$

where $c_{head}$ , $c_{base}$ and $c_{hip}$ represent the collision states of the robot's head, base and hip respectively. A state value of 1 indicates collision occurrence, while 0 indicates no collision; $\lambda_1$ , $\lambda_2$ and $\lambda_3$ are factors to balance the order of magnitude of the collision force penalties and the collision state penalties; $\mu_1$ , $\mu_2$ and $\mu_3$ are factors to adjust the penalized forces to the same order of magnitude; and $\boldsymbol{f}_{head}^{xy}$ , $\boldsymbol{f}_{base}^{xy}$ and $\boldsymbol{f}_{hip}^{xy}$ are horizontal component of the contact forces in head, base and hips, which are obtained from Isaac Lab [30].

*3) Post-Collision Velocity Reward:* The post-collision velocity reward guides the robot to move backward after collisions, prevents long-time collision with obstacles, and ensures enough time and space for pose adjustment. We reward the robot's base velocity in the direction opposite to the linear velocity command when the robot's head, base, or hip collisions happen and the robot gets stuck and cannot move along linear velocity command direction. To ensure enough time and space for pose adjustment, within a small duration after escaping, the robot's base velocity in the direction opposite to the linear velocity command is also rewarded.

$$r_{PCV} = \begin{cases} -\boldsymbol{v}_{xy} \bullet \boldsymbol{v}_{xy}^{cmd} & t_{stuck} < t_{now} < t_{escape} + \Delta t_{PCV} \\ 0 & otherwise \end{cases} \quad (10)$$

where $\boldsymbol{v}_{xy}$ and $\boldsymbol{v}_{xy}^{cmd}$ denote the robot's current linear velocity and current linear velocity command respectively, $t_{now}$ is the current time, $t_{stuck}$ is the time when the robot gets stuck and cannot move along linear velocity command direction, $t_{escape}$ is the time when the robot escapes and $\Delta t_{PCV}$ is a small duration after escaping. $\Delta t_{PCV}$ cannot be too large or too small, because too large $\Delta t_{PCV}$ causes bad velocity tracking performance, and too small $\Delta t_{PCV}$ causes an insufficient time and space for pose adjustment. In this work, we find that 0.04s is a reasonable value.

*4) Regularization Rewards:* To constrain the robot's behavior, regularization reward terms are employed. The regularization rewards include joint and body motion penalties. The joint motion penalties include penalties of action rate, action acceleration rate, joint torques, joint velocities and joint accelerations. The body motion penalties include penalties of angle velocity in *xy* in the robot base frame, linear velocity in *z* in the robot base frame, projected gravity in the robot base frame and foot slip.

## D. Training Details

Besides stairs and slopes, we design stairs tunnel and flat tunnel for training. We employ curriculum learning [11] to ensure rapid policy convergence. As shown in Fig. 4, the curriculum parameters of stairs tunnel are stair height and stairs tunnel height. The stair height is one step height of the stairs, while the stairs tunnel height is the distance between the ceiling and the stair's slope plane (a plane formed by stair upper corner line). The curriculum parameter of flat tunnel is the flat tunnel height, which is defined as the distance between the ceiling and the flat ground. As shown in Table II, we adopt a progressive terrain curriculum from simple to complex terrain. To enhance policy generalization on terrain parameters, we incorporate domain randomization in the terrain parameters including stair width, tunnel start positions, and tunnel lengths. Robot parameters' domain randomization is utilized to address the sim2real gap. The incorporated domain randomization parameters and their ranges are listed in Table III.

# IV. VERIFICATIONS

## A. Setup

We conduct policy training using the NVIDIA Isaac Sim simulator. The policy for real-world experiments is obtained through training across 2048 robots for 30,000 iterations, utilizing the curriculum learning and domain randomization settings described in Section III-D. Adam [31] method is used in networks' optimization. Training is performed on a workstation equipped with a RTX 4090D GPU, 32GB RAM, and an i7-14700KF CPU. Deployment for inference is implemented on the NVIDIA NX board of the UnitreeGo2 robot, with an inference frequency of 50 Hz and motor/IMU data interaction at 200 Hz. The low-level joint controller is a PD controller. The robot is 0.2m high when it crouches and 0.4m high when it stands with default joint positions. We compared the proposed method with existing proprioceptive locomotion methods [12][13] in crawl spaces. For the metrics, the success rate is used to evaluate the policy performance; the mean iteration time is used to evaluate the training computation; the t-Distributed Stochastic Neighbor Embedding (t-SNE) embeddings are used to evaluate the feature extraction network; and the norm ratio and cosine similarity are used to evaluate the terrain estimator.

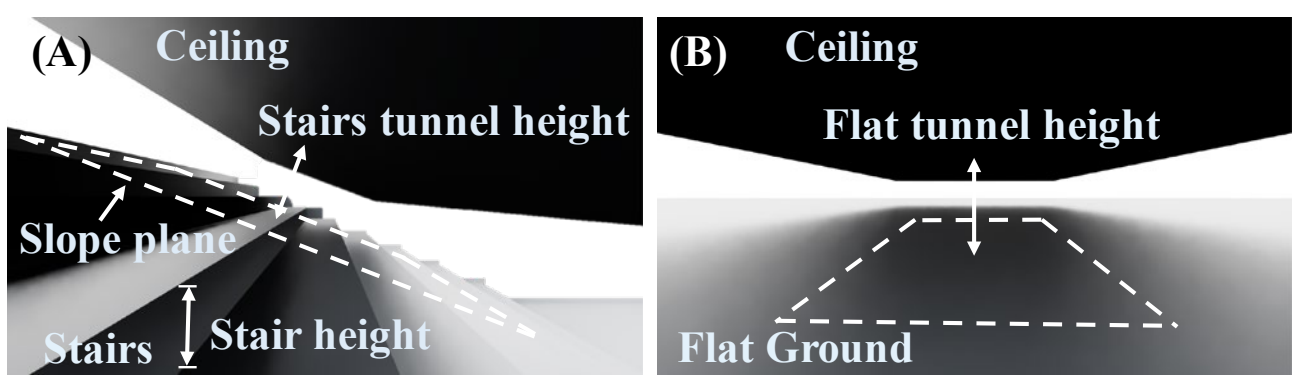

Figure 4. Two new terrain types used in our training. (A) Stairs Tunnel. The curriculum parameters are stair height and stairs tunnel height. (B) Flat Tunnel. The curriculum parameter is the flat tunnel height.

TABLE I. Reward functions

| Reward | Equation | Weight |
|---|---|---|
| **Velocity command tracking** | $\exp(-4\Vert \boldsymbol{v}_{xy} - \boldsymbol{v}_{xy}^{cmd} \Vert^2)$ | 1.0 |
| | $\exp(-4\Vert \omega_z - \omega_z^{cmd} \Vert^2)$ | 0.8 |
| **Whole-body collision penalties** | $r_{\text{collision_leg}}$ | -2.5 |
| | $r_{\text{collision_body}}$ | -2.5 |
| **Post-collision velocity** | $r_{\text{PCV}}$ | 5.0 |
| **Action rate** | $\Vert \boldsymbol{a}_t - \boldsymbol{a}_{t-1} \Vert^2$ | -0.06 |
| **Action acceleration rate** | $\Vert \boldsymbol{a}_t - 2\boldsymbol{a}_{t-1} + \boldsymbol{a}_{t-2} \Vert^2$ | -0.05 |
| **Joint torques** | $\Vert \boldsymbol{\tau} \Vert^2$ | -1e-4 |
| **Joint velocities** | $\Vert \dot{\boldsymbol{\theta}} \Vert^2$ | -6e-4 |
| **Joint accelerations** | $\Vert \ddot{\boldsymbol{\theta}} \Vert^2$ | -2e-7 |
| **Angle velocity in *xy*** | $\Vert \boldsymbol{\omega}_{xy} \Vert^2$ | -0.03 |
| **Linear velocity in *z*** | $v_z^{\ 2}$ | -3.0 |
| **Projected gravity** | $\Vert \boldsymbol{g}_{xy} \Vert^2$ | -1.0 |
| **Foot slip** | $\Vert \boldsymbol{v}_{foot} \Vert^2$ | -0.5 |

TABLE II. Terrain curriculum ranges

| Terrain Type | Ranges | Proportion |
|---|---|---|
| **Slope Up** | [0.05, 0.4]rad | 0.05 |
| **Slope Down** | [-0.05, -0.4]rad | 0.05 |
| **Stairs Up** | [0.05, 0.15]m | 0.20 |
| **Stairs Down** | [0.05, 0.15]m | 0.15 |
| **Stairs Tunnel Up** | Stairs[0.05, 0.1]m; Tunnel[0.35, 0.25]m | 0.20 |
| **Stairs Tunnel Down** | Stairs[0.05, 0.1]m; Tunnel[0.35, 0.25]m | 0.15 |
| **Flat Tunnel** | [0.4, 0.22]m | 0.20 |

TABLE III. Domain randomization ranges

| Type | Parameters | Ranges | Unit |
|---|---|---|---|
| **Hardware Randomization** | Base mass | [-1, 3] | kg |
| | Base com | [-0.01, 0.01] | m |
| | Link mass | [0.8, 1.2] × nominal value | kg |
| | Link com | [-0.005, 0.005] | m |
| | Link inertia | [0.8, 1.2] × nominal value | kg•m$^2$ |
| | Foot friction | [0.4, 1.0] | - |
| **Actuator Randomization** | Joint friction | [0, 0.001] | - |
| | Joint armature | [0, 0.001] | kg•m$^2$ |
| | Joint PD | [0.8, 1.2] × standard PD (Kp=20, Kd=0.5) | - |
| **Observation Randomization** | Time delay | [0, 0.02] | s |
| **Terrain Parameters Randomization** | Stair width | [0.25, 0.3] | m |
| | Tunnel start positions | [0.75, 1.0] | m |
| | Tunnel length | [0.8, 1.2] | m |

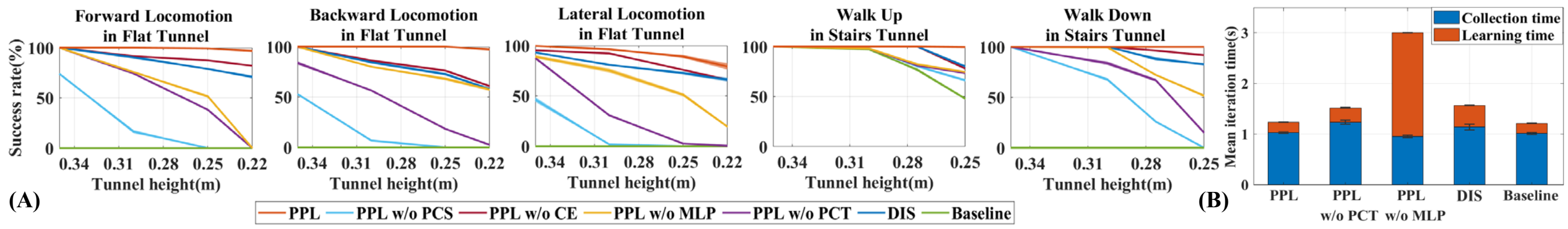

Figure 5. We compare the success rates and mean iteration time of several methods in simulations. (A) The success rates of PPL, PPL w/o PCS, PPL w/o CE, PPL w/o MLP, PPL w/o PCT, DIS and baseline in flat tunnel and stairs tunnel. (B) The mean iteration time of PPL, PPL w/o PCT, PPL w/o MLP, DIS and baseline.

- **Baseline:** Method uses elevation map as privileged environmental information and only estimates terrain information during training.
- **PPL w/o Point Cloud Supervision (PCS):** Method uses elevation maps as privileged environmental information and estimates both terrain information and collision states.
- **PPL w/o Collision Estimation (CE):** Method uses PPL-Net but only estimates environmental information, without collision estimation.
- **PPL w/o Polar Coordinate Transformation (PCT):** Method uses Cartesian coordinates and the same feature extraction method as PPL.
- **PPL w/o MLP:** Method uses Cartesian coordinates and PointNet [15] for feature extraction.
- **Depth Image Supervision (DIS):** Method uses depth image from four directions (front, rear, left, right) as privileged environmental information and simple CNN for feature extraction.
- **Quad-Traverse-Go2:** Method infers environmental information via collision domain estimation and estimates collision states. Experimental data are from the work [12].
- **MOVE:** Method uses a fixed-view camera, and considers condition without exteroceptive sensors. Experimental data are from the work [13].

### B. *Simulations*

The success rates of robot's traverse in flat tunnel and stairs tunnel are used as the performance metric. The success rates are obtained by deploying 2,000 robots and counting the number of robots that successfully traverse through flat tunnel and stairs tunnel within 20 seconds. Different random seeds are used in 10 repeated trials. We compare PPL, baseline, PPL w/o PCS, PPL w/o CE, PPL w/o PCT, PPL w/o MLP and DIS in simulations. As shown in Fig. 5A, PPL achieves the highest success rates. Baseline has no locomotion capability in flat tunnel, and has locomotion capability when walking up but no capability when walking down in stairs tunnel. PPL w/o PCS achieves higher success rates than baseline, because of the collision estimator. However, its locomotion capability is still low. PPL w/o CE achieves considerable success rates even without the collision estimator. Consequently, by further incorporating the collision estimator, PPL achieves the highest success rates. PPL achieves higher success rates than PPL w/o PCT in all cases. This is because flattening Cartesian point clouds into 1D vectors destroys the positional relationships between neighboring points, making it difficult for the network to extract meaningful features. PPL achieves higher success rates than PPL w/o MLP and DIS in all cases. This is because PPL uses a simpler network compared with PPL w/o MLP, and PPL leverages omnidirectional LiDAR scans. Using a simpler network makes PPL converge faster and have better performance in the same training data size. Using omnidirectional LiDAR scans provides a larger perception range compared with DIS. PPL achieves higher success rates for walking down than for walking up in stairs tunnel. This is because when walking down, the steps don't block movement, whereas when walking up, they become obstacles, reducing the robot's leg workspace.

Iteration time is the sum of collection time and learning time. As shown in Fig. 5B, PPL achieves fast iteration time which is comparable to baseline. Meanwhile, PPL achieves shorter learning time than PPL w/o PCT and PPL w/o MLP. This is because PPL uses representation of the point cloud in polar coordinate frame. Therefore, each point requires only one parameter for description. This representation reduces the computation of the feature extraction network. PPL and DIS have comparable collection time, and PPL requires shorter learning time. This is because the feature extraction network in PPL requires less computation than the feature extraction network in DIS.

As shown in Fig. 6B, a comparative analysis between collision states and their estimates in the simulation is performed. For different robot body parts, the estimator outputs higher collision estimates during collision and lower estimates when no collision occurs. This validates the effectiveness of our collision estimation. The absence of collisions means that the robot has exited the crawl spaces.

As shown in Fig. 6C, the temporal curve of the t-SNE embeddings for the ground and spatial feature latent representation is plotted. The two t-SNE embeddings of $z_t^l$ exhibit significant variation when the robot traverses different terrains, while remaining stable within a certain range during on the same terrain. This demonstrates that the feature extraction network effectively discriminates distinct terrain features.

As shown in Fig. 6D, the temporal curve of norm ratio ( $\left\|\tilde{z}_t^l\right\| / \left\|z_t^l\right\|$ ) and cosine similarity ( $\cos(<\tilde{z}_t^l, z_t^l>)$ ) metrics between $\tilde{z}_t^l$ and $z_t^l$ is plotted. Here $<\tilde{z}_t^l, z_t^l>$ represents the angle between $\tilde{z}_t^l$ and $z_t^l$. The closer both the norm ratio and cosine similarity are to 1, the closer the $\tilde{z}_t^l$ are to $z_t^l$. The mean norm ratio across all mentioned terrains is 0.9980 with a standard deviation of 0.0086, and the maximum deviation is 0.04. The mean cosine similarity across all mentioned terrains is 0.9977 with a standard deviation of 0.0018 and a minimum value of 0.982. It can be seen that these features are mostly equal. These indicate that the estimator effectively captures the ground and spatial features for all mentioned terrains.

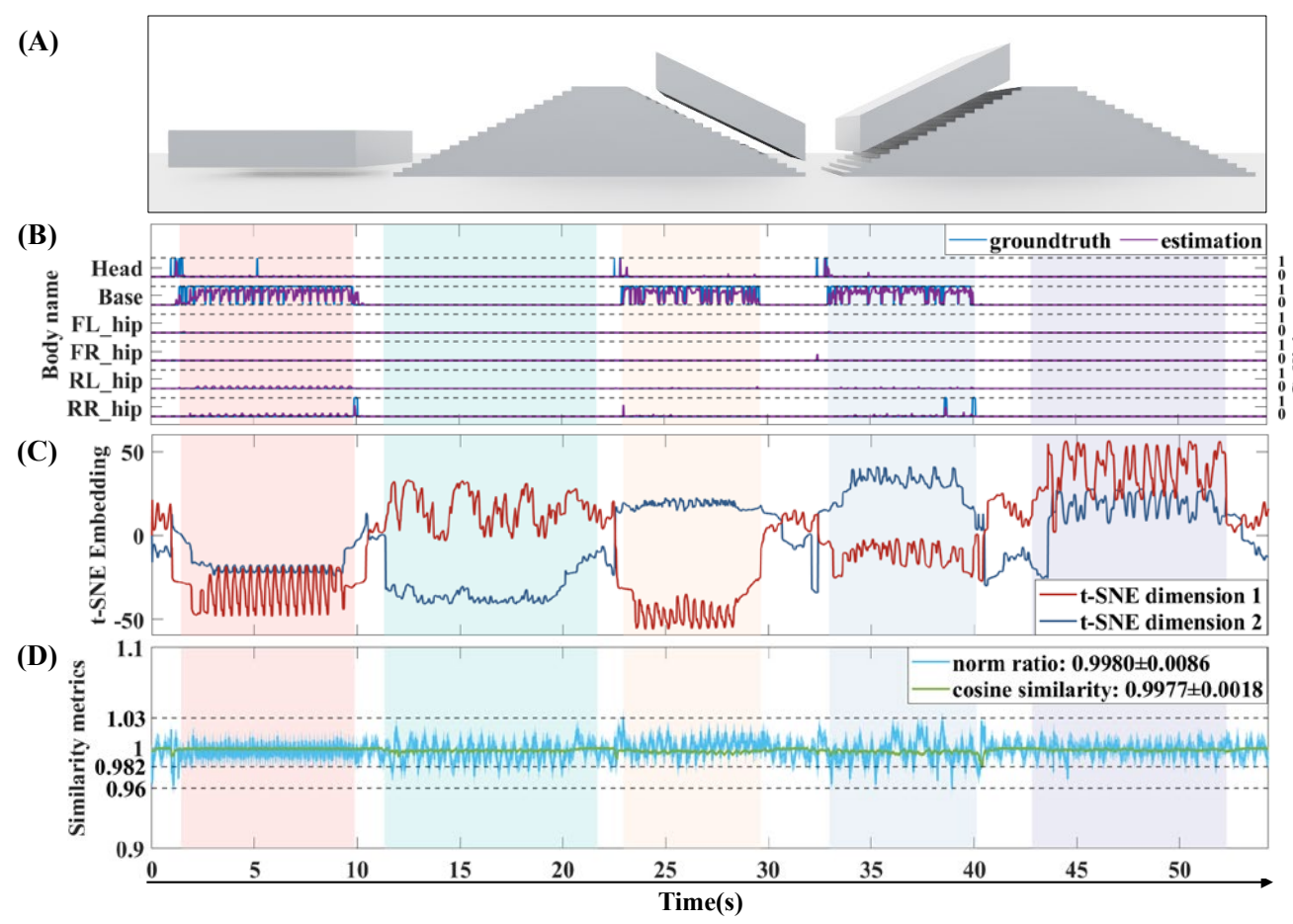

Figure 6. The robot sequentially navigates through five distinct terrains in the simulation. These terrains are indicated using red, green, yellow, blue, and purple background highlights from left to right, respectively. (A) Simulated terrain type; (B) True collision state and collision estimates; (C) t-SNE embeddings of ground and spatial features latent represents; (D) Norm ratio and cosine similarity between latent estimates and their true values.

## C. *Experiments*

We deploy the policy on the real robot to conduct experiments for testing the performance of our method. Fig. 7 presents locomotion sequences in flat tunnel and stairs tunnel. Note that the robot walks in a normal height (0.4m) outside the tunnel.

As shown in Fig. 7A, B and C, our method can traverse through 0.22m-height flat tunnel in forward, backward and lateral directions, respectively. In our experience, a 0.22m-height flat tunnel is difficult to traverse for UnitreeGo2 robot since the robot is 0.2m high when it crouches. Experiments show that the robot has effective locomotion behaviors in flat tunnel. For forward locomotion, the robot's head first lowers and enters the tunnel, followed by lowering the tail height, enabling the entire robot to enter. For backward locomotion, the tail first lowers and enters the tunnel, followed by lowering the head height, enabling the entire robot to enter. For lateral locomotion, both the head and tail are lowered simultaneously to a lower height, enabling the entire robot to enter the tunnel.

As shown in Fig. 7D and E, our method can traverse up and down through 0.25m-height stairs tunnel in forward locomotion, where the stairs' step height is 0.1m and step width is 0.25m. A slightly higher tunnel height than in flat tunnel is required for robot passage because low tunnel height on stairs reduces the robot's leg workspace, therefore, a tunnel height of 0.25 m is selected for the stairs tunnel experiment. Experiments show that the robot has learned to simultaneously handle both tunnel and stair terrains. For walking up, the robot adjusts foot height to adapt to the step ascent first, and then lowers base height and enters the tunnel; for walking down, the robot adjusts foot height to adapt to the step descent first, and then lowers its head to a lower height and enters the tunnel, followed by lowering the tail height, enabling the entire robot to enter.

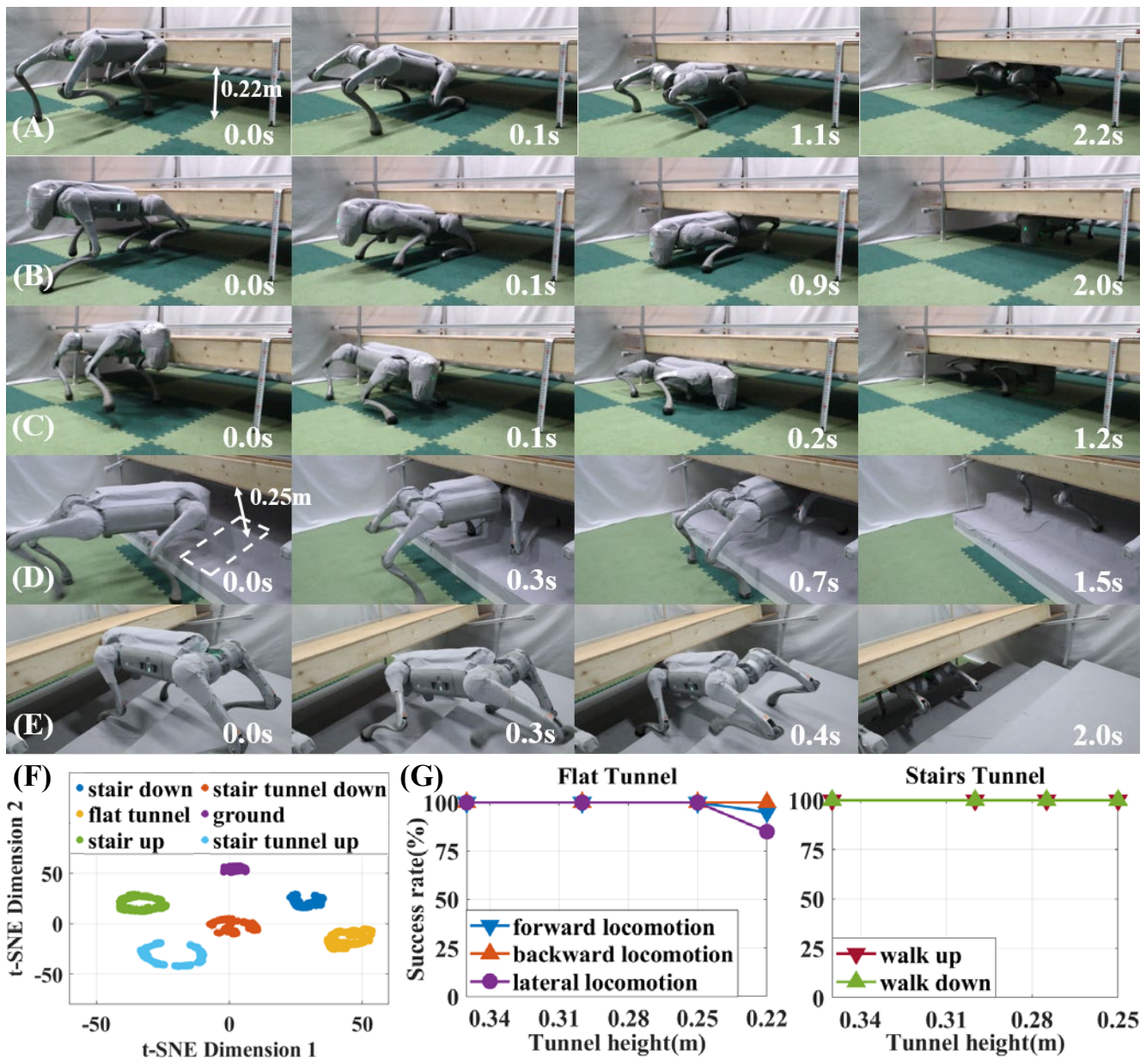

Figure 7. Locomotion image sequences in indoor experiments. (A)-(C) Locomotion in 0.22m-height flat tunnel. (D)-(E) Locomotion in 0.25m-height stairs tunnel with 0.1m-height and 0.25m-width step. (F) t-SNE analysis·for $\tilde{z}_t^l$. (G) Success rates for each case.

As shown in Fig. 7F, the 2D t-SNE visualization of $\tilde{z}_t^l$ is used to demonstrate the validity of the estimates. $\tilde{z}_t^l$ in different terrains are distinctly separated, while those in the same terrain cluster in the same regions. This demonstrates that PPL's estimation is effective in the real world.

As shown in Fig. 7G, we conduct 20 trials per terrain and calculate the real-world success rates. For flat tunnel, the robot achieves a 100% success rate in several heights of tunnel, and achieves a high success rate in the extreme low height. For stairs tunnel, the robot achieves 100% success rate in all tunnel heights. The failure cases occur during the initial entry into flat tunnel. At this time, the robot's actual adjusted body height is insufficient, resulting in entrapment. This failure arises because the robot's body thickness is 0.20 m, and the clearance for both body adjustment and leg motion are very small under the extreme low height of tunnel. Locomotion inaccuracies in the achieved body height ultimately prevent successful entry into the crawl space in this extreme low height of tunnel.

Table IV compares our method with existing proprioceptive locomotion methods in crawl spaces. To eliminate the effect of robot size, we adopt relative height (the ratio of tunnel height to the robot height in crouching state) as a performance metric. A smaller relative height indicates a more challenging environment for the robot. As shown in Table IV, our method makes the robot crawl through the lowest flat tunnel without requiring exteroceptive sensors. Moreover, to the best of our knowledge, our method makes the robot crawl through stairs tunnel for the first time.

TABLE IV. Comparisons with existing methods on sensor dependence and crawled tunnel relative height

| Method | Exteroceptive Sensor Dependence | Flat Tunnel Relative Height | Stairs Tunnel Relative Height |
|---|---|---|---|
| **Quad-Traverse-Go2[12]** | Without exteroceptive sensor | 1.35(Forward) | unknown |
| **MOVE[13]** | Without exteroceptive sensor | 1.56 (Forward, Backward and Lateral) | unknown |
| | Depth camera | 1.25(Forward) | unknown |
| **PPL (Ours)** | **Without exteroceptive sensor** | **1.10 (Forward, Backward and Lateral)** | **1.25 at 0.1m stairs** |

We also conduct experiments in other kinds of crawl spaces as shown in Fig. 1. The results show that the method is applicable in these conditions. The experimental video is provided in the supplemental material.

## V. Conclusion

Legged locomotion in crawl spaces is challenging. In this study, a point cloud supervised proprioceptive locomotion reinforcement learning method for legged robots in crawl spaces is proposed. A state estimation network is designed to estimate the robot's collision states and infer ground and spatial features from historical proprioceptive data. The true ground and spatial features are extracted efficiently using the point cloud representation in the polar coordinate frame, and are used to supervise the state estimation network learning. Experiments in flat tunnel and stairs tunnel are conducted to validate the effectiveness of the proposed method. In the future, some tactile sensors can be put on the robot surface, and the PPL can be extended to a tactile sense based proprioceptive locomotion. The PPL-Net can also be used for hybrid exteroceptive and proprioceptive locomotion of legged robots in 3D environments.